\documentclass[10pt,twocolumn,letterpaper]{article}

\usepackage{wacv}              
\usepackage[accsupp]{axessibility}

\usepackage{adjustbox}

\definecolor{wacvblue}{rgb}{0.21,0.49,0.74}
\usepackage[pagebackref,breaklinks,colorlinks,allcolors=wacvblue]{hyperref}

\def\wacvPaperID{*****} 
\def\confName{WACV}
\def\confYear{2026}

\title{ReaMIL: Reasoning- and Evidence-Aware Multiple Instance Learning for Whole-Slide Histopathology}

\author{
Hyun Do Jung$^{1}$ \qquad
Jungwon Choi$^{2}$ \qquad
Hwiyoung Kim$^{1}$\thanks{Corresponding author.}\\
$^{1}$Yonsei University \qquad $^{2}$KAIST
}

\begin{document}
\maketitle
\begin{abstract}
We introduce ReaMIL (Reasoning- and Evidence-Aware MIL), a multiple instance learning approach for whole-slide histopathology that adds a light selection head to a strong MIL backbone. The head produces soft per-tile gates and is trained with a budgeted-sufficiency objective: a hinge loss that enforces the true-class probability to be $\geq \tau$ using only the kept evidence, under a sparsity budget on the number of selected tiles. The budgeted-sufficiency objective yields small, spatially compact evidence sets without sacrificing baseline performance. Across TCGA-NSCLC (LUAD vs.\ LUSC), TCGA-BRCA (IDC vs.\ Others), and PANDA, ReaMIL matches or slightly improves baseline AUC and provides quantitative evidence-efficiency diagnostics. On NSCLC, it attains AUC 0.983 with a mean minimal sufficient K (MSK) $\approx 8.2$ tiles at $\tau = 0.90$ and AUKC $\approx 0.864$, showing that class confidence rises sharply and stabilizes once a small set of tiles is kept. The method requires no extra supervision, integrates seamlessly with standard MIL training, and naturally yields slide-level overlays. We report accuracy alongside MSK, AUKC, and contiguity for rigorous evaluation of model behavior on WSIs.
\end{abstract}
    
\section{Introduction}
\label{sec:Introduction}

Whole-slide histopathology has become a standard testbed for weakly supervised learning~\cite{dimitriou2019deep}. Modern scanners produce gigapixel slides, but in most clinical datasets only slide-level labels are available: tumor subtype, grade, or outcome, without any pixel- or patch-level annotations~\cite{litjens2018camelyon}. Multiple instance learning (MIL) provides a natural framework for this setting, treating each slide as a bag of tiles that are encoded and aggregated into a single prediction~\cite{campanella2019clinical, ilse2018attention, lu2021clam}. Despite the weak supervision, these models can reach pathologist-level performance on some benchmarks and are now being deployed in early-stage clinical decision support tools.
\vspace{2mm}

However, standard MIL training focuses on bag-level accuracy: the model is rewarded for predicting the correct slide label, with no explicit notion of which tiles actually constitute the ``evidence'' for that prediction. Attention weights are often interpreted as explanations, but they are a side effect of training, not a primary objective~\cite{serrano2019attention, Pruthi_Gupta_Dhingra_Neubig_Lipton_2020}. This gap between slide-level performance and tile-level reasoning becomes critical when models are meant to support clinical decisions. In practice, pathologists justify diagnoses by pointing to specific regions---glands with certain architecture, nests of atypical cells, or characteristic tumor--stroma interfaces. Computational models should ideally do the same: highlight a compact set of tiles sufficient to support the predicted label, while showing that the rest of the slide does not drive the decision.

Recent advances in representation learning have shifted the landscape toward foundation models pretrained on millions of tiles across sites and organs~\cite{chen2024uni, ciga2022overcoming}. We leverage pre-extracted UNI2-h~\cite{chen2024uni} features as patch-level representations, allowing us to focus on the reasoning layer. On top of these frozen features, transformer-based MIL backbones such as TransMIL~\cite{shao2021transmil} already achieve competitive performance on multiple WSI benchmarks. Yet this ``foundation MIL'' stack does not address interpretability~\cite{Chakraborty_Tomsett_Raghavendra_Harborne_Alzantot_Cerutti_Srivastava_Preece_Julier_Rao_et_al._2017}: we have powerful encoders and backbones, but how they use evidence inside the bag remains opaque.

Our work treats evidence selection as a first-class objective in MIL rather than an afterthought. We attach a lightweight selection head on top of a strong MIL backbone to produce soft selection scores over tiles. These scores define three views of each slide: a \emph{full} bag, a \emph{keep} bag retaining only evidence tiles, and a \emph{drop} bag containing the complement. By feeding these three bags through a shared backbone, we explicitly shape how the model uses evidence through a budgeted sufficiency objective: the keep bag should reach a target confidence $\tau$ for the true class while the drop bag does not support the true label (its true-class probability remains low). We regularize evidence to be spatially compact and penalize selecting too many tiles, yielding four concrete properties: sufficiency, exclusion, contiguity, and budget. We call this framework ReaMIL: reasoning- and evidence-aware MIL.

To measure these properties, we introduce diagnostics that probe how the model's true-class probability grows as we reveal more top-scoring tiles. The area under this ``K-curve'' (AUKC) and the minimal sufficient K (MSK) at a chosen confidence threshold summarize how quickly the model's belief saturates. Across TCGA-NSCLC, TCGA-BRCA~\cite{weinstein2013cancer, edwards2015cptac}, and PANDA~\cite{bulten2022artificial}, we show that ReaMIL preserves or improves baseline AUC while substantially reducing MSK and improving AUKC, indicating that high-confidence decisions can be supported by small, spatially compact sets of tiles.

In summary, the main contributions of this work are summarized as follows:

\begin{itemize}
    \item We present ReaMIL, a reasoning- and evidence-aware MIL framework that integrates sufficiency, exclusion, spatial contiguity, and evidence sparsity.
    
    \item We introduce quantitative evidence-efficiency metrics, including minimal sufficient~$K$ (MSK) and the area under the K-curve (AUKC), which measure how quickly confidence emerges as diagnostic tiles are revealed.
    
    \item We demonstrate that our ReaMIL maintains or even improves slide-level performance while producing  highly compact and spatially coherent evidence sets across TCGA-NSCLC, TCGA-BRCA, and PANDA.
\end{itemize}

\section{Related Work}
\label{sec:Related Works}

\subsection{Multiple instance learning for whole-slide histopathology}

Multiple instance learning (MIL) treats a digital slide as a bag of tiles with a single slide-level label and no supervision for individual tiles. Attention-based pooling, introduced by Ilse et al.~\cite{ilse2018attention}, replaced fixed max- or mean-pooling with a learned weighted combination of tile features and became the standard aggregation strategy. Subsequent architectures incorporated class-specific attention and clustering constraints (CLAM~\cite{lu2021clam}) or transformer-based self-attention to model long-range context between tiles (TransMIL~\cite{shao2021transmil}). More recently, feature extraction has been decoupled from MIL aggregation: large self-supervised or multimodal encoders pre-trained on millions of histology tiles are frozen, and MIL models operate on pre-extracted features~\cite{chen2022scaling}. This reduces training cost and improves robustness across cohorts. We follow this strategy, using UNI2-h as the feature backbone while the MIL component focuses on aggregating and selecting evidence at the tile level.

\subsection{Interpretability of attention in MIL}

Interpretability in MIL for pathology has largely relied on visualizing attention weights as heatmaps or displaying top-attended tiles~\cite{ilse2018attention, lu2021clam}. However, attention as explanation has well-known limitations~\cite{serrano2019attention}: attention scores are shaped by end-to-end training and may not reflect causal importance; high-attention tiles can be redundant or partially spurious; and there is no guarantee that the attended subset alone suffices to recover the correct prediction, nor that the complement is non-predictive. Various remedies—instance-level regularization, auxiliary classifiers, multiple attention heads, or region proposals from slide labels—can make heatmaps more visually convincing, but they typically lack a quantitative framework for measuring how much evidence is actually needed for a decision.

\subsection{Budgeted evidence and selective prediction}

The idea of constraining a model to rely on a small subset of inputs appears in selective prediction~\cite{geifman2017selective}, budgeted or early-exit models, and rationalization methods that train differentiable selectors to pick a few tokens or patches so that a downstream predictor matches the full model using only the selected subset. Our work adapts this perspective to MIL: we attach a small selection head on top of a fixed MIL encoder and train it with losses enforcing sufficiency of the kept bag, exclusion of the dropped bag, spatial contiguity, and an explicit budget on selection rate (normalized selection mass). We quantify the resulting behavior with K-curves, minimal sufficient $K$ (MSK), and area under the K-curve (AUKC)—metrics that capture how quickly confidence rises as diagnostic regions are added and how small a subset suffices for a diagnosis.

\section{Methodology}
\label{sec:Methodology}

\begin{figure*}[t]
 \centering
   \includegraphics[width=\textwidth]{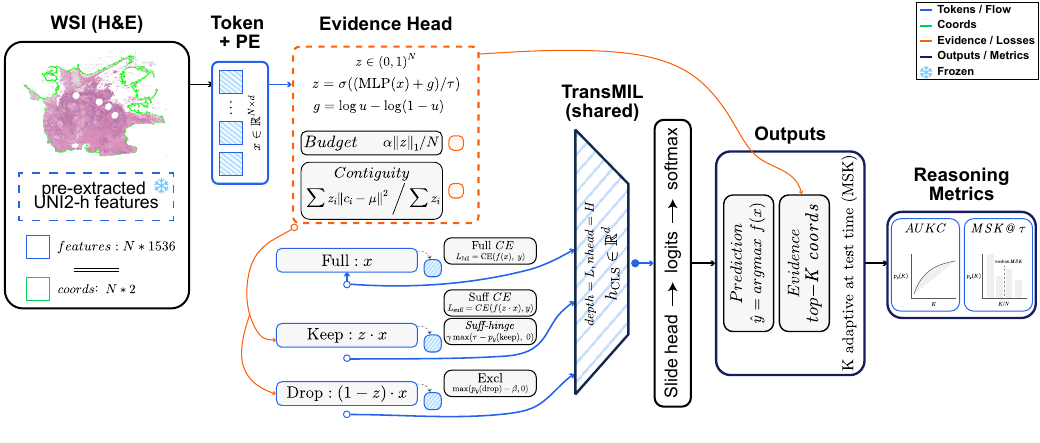}
\vspace{-2em}
   \caption{Overview of ReaMIL. Frozen UNI2-h features and patch coordinates are extracted from each WSI and mapped to tokens with positional embeddings. An evidence head produces soft selection scores \(z \in (0,1)^N\) via a Concrete (Gumbel--sigmoid) gate, and defines three bags: the full bag \(x\), a keep bag \(z \cdot x\), and a drop bag \((1-z) \cdot x\). All three bags are processed by a shared TransMIL encoder and slide head. Losses encourage (i) correct predictions on the full and keep bags (cross-entropy on \(\ell_{\text{full}}\) and \(\ell_{\text{keep}}\) plus a sufficiency hinge at confidence \(\tau\)), (ii) low true-class probability on the drop bag (exclusion), (iii) spatially compact selections (contiguity on coordinates), and (iv) a small evidence budget via an \(\ell_1\) penalty on \(z\). At test time, the model outputs both slide predictions and ranked evidence coordinates. Reasoning metrics are computed by probing the top-K curve of true-class probability \(p_y(K)\): AUKC summarizes the area under this curve, and \(\mathrm{MSK}@\tau\) measures the minimal number of tiles required to reach confidence \(\tau\).
}
   \label{fig:pipeline}
\end{figure*}

We build ReaMIL on top of a transformer-based MIL backbone, adding a lightweight evidence head that learns which patches suffice for the slide-level prediction. Figure~\ref{fig:pipeline} illustrates the overall architecture.

\subsection{Problem setup and backbone}

Following standard weakly supervised MIL, each slide $s$ consists of a bag of patch features $X_s = \{x_{s,i}\}_{i=1}^{N_s}$ extracted by a frozen encoder, along with spatial coordinates $C_s = \{c_{s,i}\}_{i=1}^{N_s}$ where $c_{s,i} = (u_{s,i}, v_{s,i})$ is the pixel location of patch $i$. We use UNI2-h~\cite{chen2024uni} to extract $d{=}1536$ dimensional features. The slide has a single label $y_s \in \{1, \ldots, C\}$ but no patch-level supervision.

Patch features are projected into a token space via $\tilde{x}_{s,i} = W_{\text{feat}} x_{s,i} + b_{\text{feat}}$, with optional positional embeddings $t_{s,i} = \tilde{x}_{s,i} + \text{MLP}_{\text{pos}}(\text{norm}(c_{s,i}))$. The resulting tokens $T_s = [t_{s,1}, \ldots, t_{s,N_s}]$ are processed by a TransMIL backbone~\cite{shao2021transmil}: a learned [CLS] token is prepended to the sequence and passed through $L$ transformer layers. The final CLS representation $h_{\text{CLS}} \in \mathbb{R}^{d_{\text{model}}}$ is mapped to class logits $\ell_s = W_{\text{cls}} h_{\text{CLS}} + b_{\text{cls}} \in \mathbb{R}^C$, and baseline training uses cross-entropy $\mathcal{L}_{\text{full}} = \text{CE}(\ell_s, y_s)$.

\subsection{Evidence selection head}

For each token $t_{s,i}$, a small MLP computes a selection logit $a_{s,i} = \text{MLP}_{\text{sel}}(t_{s,i}) \in \mathbb{R}$. To enable differentiable selection, we apply the Concrete (Gumbel-sigmoid) relaxation~\cite{maddison2017concrete, jang2017categorical}. We sample $\epsilon_{s,i} \sim \text{Uniform}(0,1)$ and compute:
\begin{equation}
z_{s,i} = \sigma\!\left(\frac{a_{s,i} + \log \epsilon_{s,i} - \log(1-\epsilon_{s,i})}{T}\right) \label{eq:1}
\end{equation}
where $T > 0$ is the temperature. This yields soft selection scores $z_{s,i} \in (0,1)$ that approach binary values as $T \to 0$.

The scores define three views of each slide: the original bag $X_{\text{full}} = X_s$, the evidence bag $X_{\text{keep}} = z_s \odot X_s$, and its complement $X_{\text{drop}} = (1-z_s) \odot X_s$, where $\odot$ denotes element-wise scaling. Since hard selection is non-differentiable, we retain all tokens in the sequence but down-weight non-selected patches via soft masking. Each view is processed by the shared backbone to produce logits $\ell_{\text{full}}$, $\ell_{\text{keep}}$, and $\ell_{\text{drop}}$.

\subsection{Evidence-aware training objectives}

Our goal is not only to achieve high slide-level accuracy, but also to explicitly shape how the model uses evidence inside each bag. To this end, we design an evidence-aware training objective that couples a standard classification loss with four additional terms, each enforcing a distinct property of the selector. Together, these losses encourage decisions that are (i) \emph{sufficient}, with a small subset of selected patches supporting high-confidence predictions; (ii) \emph{exclusive}, with the remaining patches not supporting the true label (low true-class probability); (iii) \emph{spatially contiguous}, so that evidence forms coherent regions on the slide; and (iv) \emph{budgeted}, limiting the amount of selected evidence.

Let $p_y(\ell) = \mathrm{softmax}(\ell)[y_s]$ denote the true-class probability. We combine five losses:
\begin{align}
    \mathcal{L}_{\text{full}}   &= \mathrm{CE}(\ell_{\text{full}}, y_s), \\
    \mathcal{L}_{\text{suff}}   &= \mathrm{CE}(\ell_{\text{keep}}, y_s)
        + \max\big(\tau - p_y(\ell_{\text{keep}}), 0\big), \\
    \mathcal{L}_{\text{excl}}   &= \max\big(p_y(\ell_{\text{drop}}) - \beta, 0\big), \\
    \mathcal{L}_{\text{contig}} &= 
        \frac{\sum_i z_{s,i} \,\lVert c_{s,i} - \mu_s \rVert_2^2}
             {\sum_i z_{s,i}}, \\
    \mathcal{L}_{\text{budget}} &= \frac{1}{N_s}\sum_i z_{s,i},
\end{align}
where $z_{s,i}$ are selection scores, 
$\mu_s = \sum_i z_{s,i} c_{s,i} / \sum_i z_{s,i}$ is the $z$-weighted centroid,
and $\tau, \beta \in (0,1)$ are hyperparameters, with $\tau$ used as a confidence threshold on the true-class probability $p_y(.)$ and $T$ in \eqref{eq:1} serving as the temperature of the Concrete gate. $\mathcal{L}_{\text{budget}}$ is the average selection rate (normalized $\ell_1$ norm of $z_s$) and acts as an explicit sparsity penalty.

The total loss is
\begin{equation}
    \mathcal{L} = \mathcal{L}_{\text{full}}
      + \lambda_{\text{suff}} \mathcal{L}_{\text{suff}}
      + \lambda_{\text{excl}} \mathcal{L}_{\text{excl}}
      + \lambda_{\text{contig}} \mathcal{L}_{\text{contig}}
      + \lambda_{\text{budget}} \mathcal{L}_{\text{budget}}.
\end{equation}
Here, the weights $\lambda_{\text{suff}}, \lambda_{\text{excl}}, \lambda_{\text{contig}}, \lambda_{\text{budget}}$ balance fidelity against the strength of the evidence-aware constraints.

\subsection{Evidence-efficiency metrics}

Conventional metrics such as AUC, accuracy, or F1 summarize how often a model predicts the correct slide label, but they are insensitive to \emph{how much} evidence the model needs to make those predictions. To evaluate whether ReaMIL actually learns to rely on small, sufficient evidence sets, we introduce a family of evidence-efficiency metrics based on the behavior of the model as top-ranked tiles are gradually revealed. To quantify evidence efficiency, we probe the relationship between revealed patches and model confidence. At test time, we rank patches by their selection logits $a_{s,i}$ (Gumbel noise is used only during training) and construct a \textbf{K-curve} that records the true-class probability $p_y(K)$ as a function of the number of revealed patches $K$.

    
\paragraph{Minimal Sufficient K (MSK).}
For each slide $s$ and confidence threshold $\tau$, we define
\begin{equation}
    \mathrm{MSK}_s(\tau) = \min \{ K : p_y(K) \ge \tau \}.
\end{equation}
MSK measures how many top-ranked patches are needed for the model to reach confidence $\tau$.

\paragraph{Area Under K-Curve (AUKC).}
We also define the area under the K-curve in terms of the normalized evidence fraction $\kappa = K / N_s \in [0,1]$:
\begin{equation}
    \mathrm{AUKC}_s = \int_0^1 p_y(\kappa)\, d\kappa,
\end{equation}
where $p_y(\kappa)$ denotes the true-class probability when the top $\kappa \cdot N_s$ tiles are kept.
\section{Experiments}
\label{sec:Experiments}

\begin{figure}[t]
  \centering
  \includegraphics[width=\linewidth]{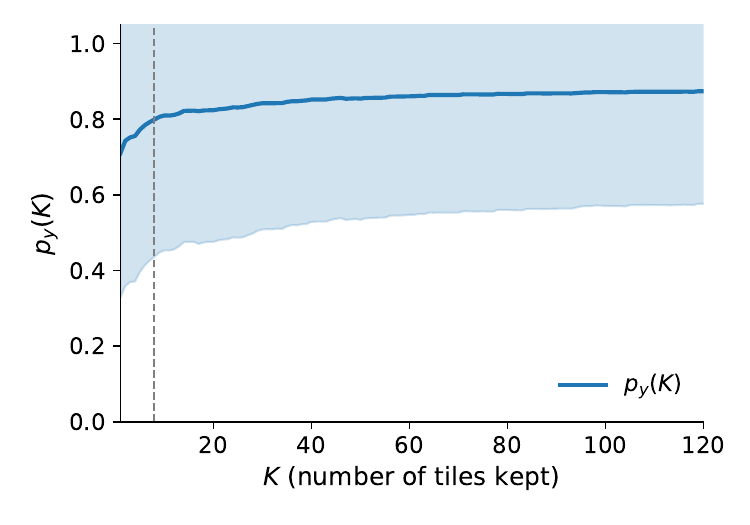}
\vspace{-2em}
  \caption{K-curve on NSCLC (test set). True-class probability $p_y(K)$ as top-$K$ tiles (ranked by selection score) are revealed. Solid line: mean across slides; shaded region: $\pm$1 std. Vertical dashed line: mean MSK@$\tau=0.9$. MSK is computed per-slide before averaging, so individual slides may cross $\tau$ even when the mean curve does not.}
  \label{fig:kcurves}
  \vspace{-4mm}
\end{figure}

\begin{figure*}[!t]
  \centering
  \includegraphics[width=0.48\textwidth]{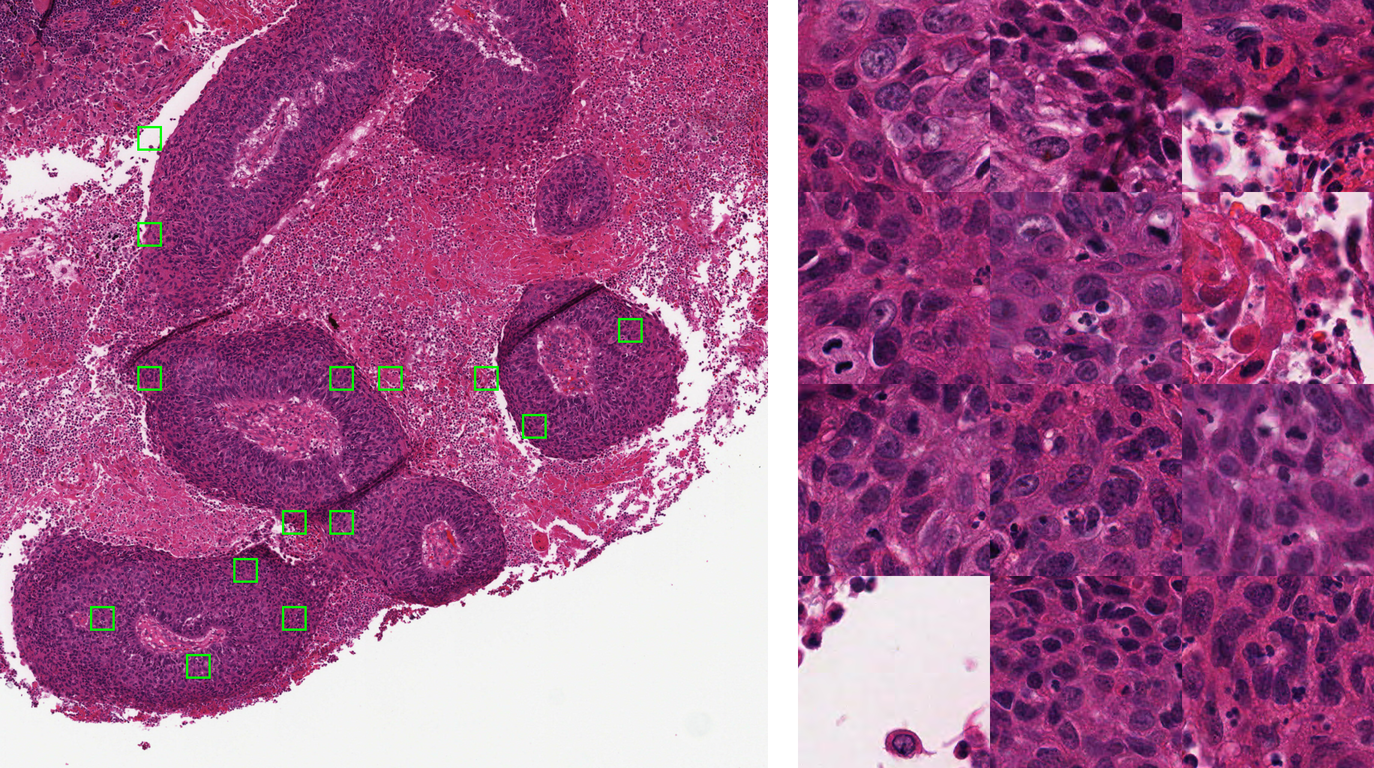}
  \hfill
  \includegraphics[width=0.48\textwidth]{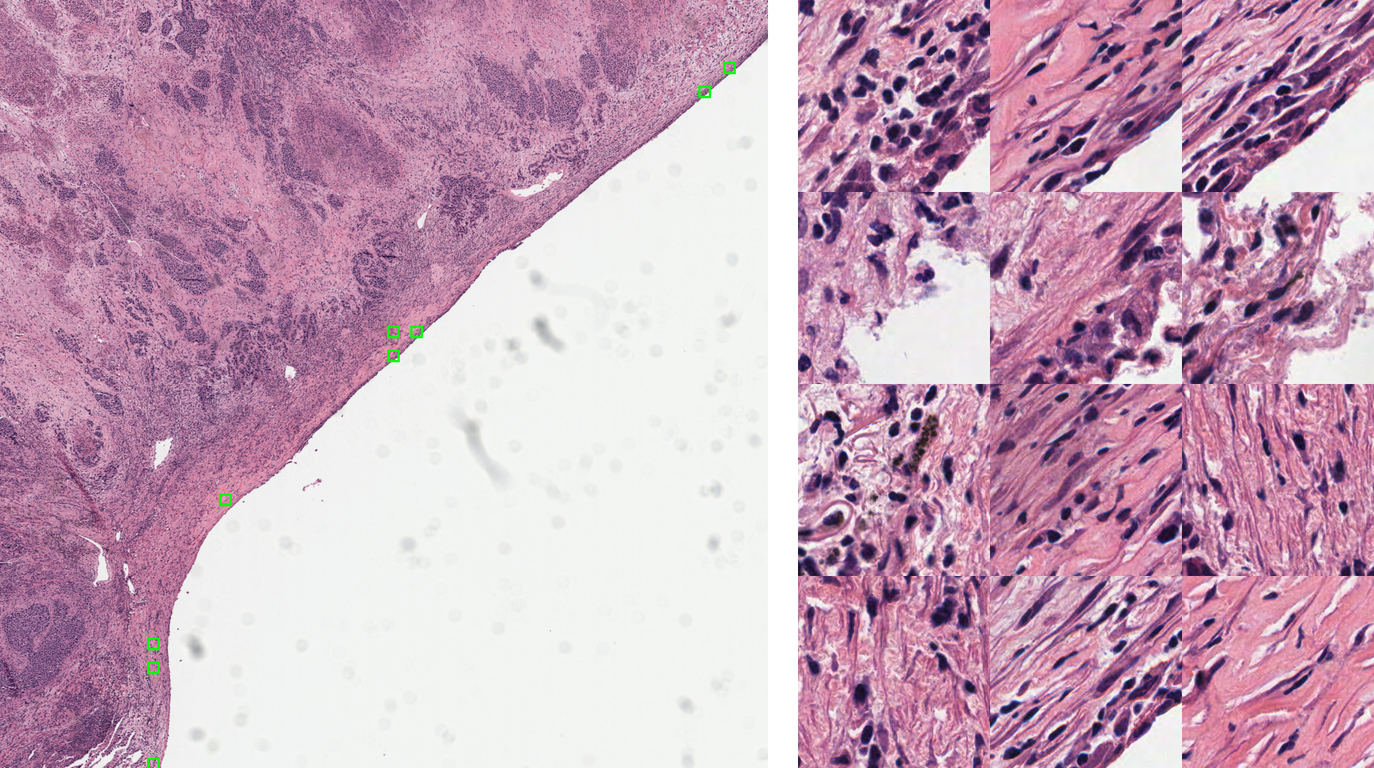}
  \caption{Evidence visualization on TCGA-NSCLC.  \textbf{Left:} LUSC (squamous cell carcinoma) case with relatively compact evidence clusters over squamous tumor nests. 
  \textbf{Right:} LUAD (adenocarcinoma) case with more diffuse selection over gland-forming tumor regions. 
  Each panel shows selected tile locations (green boxes) and the corresponding top-$K$ zoomed patches. 
  For visualization, we show zoomed-in regions (left: $8192 \times 8192$; right: $16384 \times 16384$ pixels), where the selected tiles (size $256 \times 256$) are outlined in green.}
  \label{fig:overlays}
  \vspace{-3mm}
\end{figure*}

\subsection{Datasets and setup}

We evaluate on three binary WSI classification tasks: \textbf{TCGA-NSCLC} (LUAD vs.\ LUSC), \textbf{TCGA-BRCA} (IDC vs.\ Others), and \textbf{PANDA} (clinically significant vs.\ non-significant prostate cancer). For each dataset, we construct patient-disjoint train/validation/test splits with class stratification. All slides are processed into frozen UNI2-h features ($d{=}1536$) and tile coordinates; the encoder is never fine-tuned. Details on label mappings and patient counts are in the supplement.

The backbone is a TransMIL-style transformer ($d_{\text{model}}{=}512$, 8 heads, 4 layers). We first train a baseline with standard cross-entropy, then attach the evidence head and continue training with the combined loss (Section~\ref{sec:Methodology}), warm-starting from the baseline checkpoint. All models use AdamW with cosine decay and mixed-precision on two RTX 6000 Ada GPUs. We report mean$\pm$std over three seeds.

\subsection{Slide-level performance}

Table~\ref{tab:main} compares the baseline (TransMIL + UNI2-h, no evidence head) against ReaMIL with the full budgeted objective, showing that adding the evidence head and reasoning losses extends standard MIL pipelines without trading accuracy for interpretability.

\begin{table}[h]
\centering
\small
\setlength{\tabcolsep}{2.5pt}
\begin{tabular}{llccc}
\toprule
Dataset & Method & AUC & Acc & F1$_{\text{macro}}$ \\
\midrule
BRCA  & Baseline & 0.897$\pm$0.019 & 0.877$\pm$0.006 & 0.819$\pm$0.022 \\
      & $+$ReaMIL   & 0.904$\pm$0.011 & 0.888$\pm$0.010 & 0.827$\pm$0.019 \\
\midrule
NSCLC & Baseline & 0.969$\pm$0.006 & 0.935$\pm$0.006 & 0.935$\pm$0.006 \\
      & $+$ReaMIL   & 0.983$\pm$0.004 & 0.927$\pm$0.025 & 0.927$\pm$0.026 \\
\midrule
PANDA & Baseline & 0.985$\pm$0.002 & 0.955$\pm$0.004 & 0.945$\pm$0.004 \\
      & $+$ReaMIL   & 0.989$\pm$0.003 & 0.958$\pm$0.002 & 0.948$\pm$0.003 \\
\bottomrule
\end{tabular}
\caption{Slide-level performance (mean$\pm$std, 3 seeds). ReaMIL uses the full budgeted objective.}
\label{tab:main}
\end{table}

\subsection{Evidence efficiency}

Figure~\ref{fig:kcurves} shows K-curves on NSCLC: for each slide, tiles are ranked by selection score and the true-class probability $p_y(K)$ is recorded as the top-$K$ tiles are revealed. Table~\ref{tab:reasoning} reports MSK@$\tau{=}0.90$ (minimal tiles to reach 90\% confidence) and AUKC across all datasets. Note that these metrics require an explicit selector to rank tiles and are therefore defined only for ReaMIL, not for vanilla MIL baselines.

\begin{table}[ht]
\centering
\small
\begin{tabular}{lcc}
\toprule
Dataset & MSK@$\tau{=}0.90$ ($\downarrow$) & AUKC ($\uparrow$) \\
\midrule
BRCA   & 16.0$\pm$11.8 & 0.833$\pm$0.018 \\
NSCLC  & 8.2$\pm$2.1 & 0.864$\pm$0.069 \\
PANDA  & 7.2$\pm$3.6 & 0.811$\pm$0.055 \\
\bottomrule
\end{tabular}
\caption{Evidence efficiency metrics for ReaMIL (mean$\pm$std, 3 seeds). MSK@0.9: minimal tiles to reach 90\% confidence. AUKC: area under the K-curve. These metrics require an explicit selector and are not defined for vanilla MIL baselines.}
\label{tab:reasoning}
\vspace{-3mm}
\end{table}

On NSCLC, ReaMIL achieves MSK@0.9 of approximately 8.2 tiles—fewer than 0.1\% of the average bag size (${\sim}$6,000 tiles)—demonstrating that the selector concentrates evidence into a small, sufficient subset.

\subsection{Ablations}

Table~\ref{tab:ablation} isolates each loss component on NSCLC. \mbox{Without} the full objective, ablated models select nearly all tiles (mean $\|z\|_1 > 0.85$ vs.\ $0.002$ for ReaMIL), causing the keep bag to approximate the full bag. This yields trivially low suff.\
gap and contig.\ values—not because evidence is well-selected, but because almost nothing is excluded. In contrast, ReaMIL (full) achieves true sparse selection: $p_y(\text{drop}) \approx 0$ shows the complement is non-predictive for the true class, confirming that the small selected set genuinely captures the diagnostic signal. 

\begin{table}[h]
\centering
\small
\setlength{\tabcolsep}{1pt}
\begin{adjustbox}{width=1.0\linewidth, totalheight=\textheight, keepaspectratio}
\begin{tabular}{lccccc}
\toprule
Variant & AUC & Suff. gap ($\downarrow$) & $p_y$(drop) ($\downarrow$) & Contig. ($\downarrow$) & $\|z\|_1$ ($\downarrow$) \\
\midrule
ReaMIL (full)     & 0.984 & 0.119 & 0.000 & 0.137 & \textbf{0.002} \\
\quad w/o sufficiency   & 0.981 & 0.039 & 0.167 & 0.106 & 0.847 \\
\quad w/o exclusion     & 0.981 & 0.000 & 0.414 & 0.128 & 0.923 \\
\quad w/o contiguity    & 0.985 & 0.001 & 0.339 & 0.127 & 0.891 \\
\bottomrule
\end{tabular}
\end{adjustbox}
\caption{Ablations on NSCLC. Suff.\ gap: confidence drop using only kept tiles. $p_y$(drop): true-class probability of the drop bag (lower = the drop bag alone does not support the true label). Contig.: spatial dispersion. $\|z\|_1$: mean selection rate (normalized $\ell_1$; lower = sparser). Ablations select nearly all tiles, yielding trivially low suff.\ gap but defeating the goal of compact evidence; only ReaMIL (full) achieves true sparse selection.}
\label{tab:ablation}
\vspace{-3mm}
\end{table}
\subsection{Qualitative results}

Figure~\ref{fig:overlays} shows evidence overlays on representative NSCLC slides. 
The LUSC case (left) exhibits relatively compact evidence clusters over squamous tumor nests. 
The LUAD case (right) shows a more diffuse pattern of selected tiles across gland-forming adenocarcinoma regions. 
In both cases, ReaMIL concentrates its evidence on morphologically relevant tumor areas while largely ignoring background tissue, consistent with the quantitative findings. 
\section{Conclusion}
\label{sec:Conclusion}

We presented ReaMIL, a method that transforms whole-slide classification into an evidence-seeking problem by adding a budgeted selection head to standard MIL backbones. Training the selector so that a small, spatially compact subset suffices for prediction while forcing complementary tiles to be non-predictive for the true class preserves baseline AUC while producing compact evidence---on TCGA-NSCLC, AUC 0.983 with MSK $\approx 8.2$ at $\tau = 0.90$ and AUKC $\approx 0.864$. The framework requires only slide-level supervision, fits existing pipelines, and shows that accurate yet interpretable MIL is achievable without extra annotation---critical as computational pathology moves toward clinical deployment.

\noindent\textbf{Limitations.} Our approach relies on pre-extracted features from a single foundation model (UNI2-h) and has been evaluated on relatively balanced research datasets. Validation on more diverse clinical cohorts with class imbalance and domain shift, as well as user studies with pathologists to assess clinical utility, remain important directions for future work.

\section*{Acknowledgement}
This work was supported by the Bio-industrial Technology Development Program (RS-2025-02220286, (Division 2) Development of large language AI model-based techniques and platforms for nursery record generation and task automation) funded by the Ministry of Trade, Industry \& Resources (MOTIR, Korea)

{
    \small
    \bibliographystyle{ieeenat_fullname}
    \bibliography{main}
}

\end{document}